\documentclass[
]{ceurart}

\usepackage{listings}
\usepackage{comment}
\usepackage{graphicx}
\usepackage{hyperref}

\usepackage{todonotes}

\usepackage{eso-pic}
\usepackage{graphicx}

\begin{document}

\copyrightyear{2025}
\copyrightclause{Copyright for this paper by its authors.
  Use permitted under Creative Commons License Attribution 4.0
  International (CC BY 4.0).}

\conference{OM 2025: The 20th International Workshop on Ontology Matching collocated with the 24nd International Semantic Web Conference ISWC-2025 November 02-06, 2025, Nara, Japan}
\makeatletter
\def\insertlogo{%
  \begin{picture}(0,0)
    \put(350, 20){\includegraphics[width=5cm]{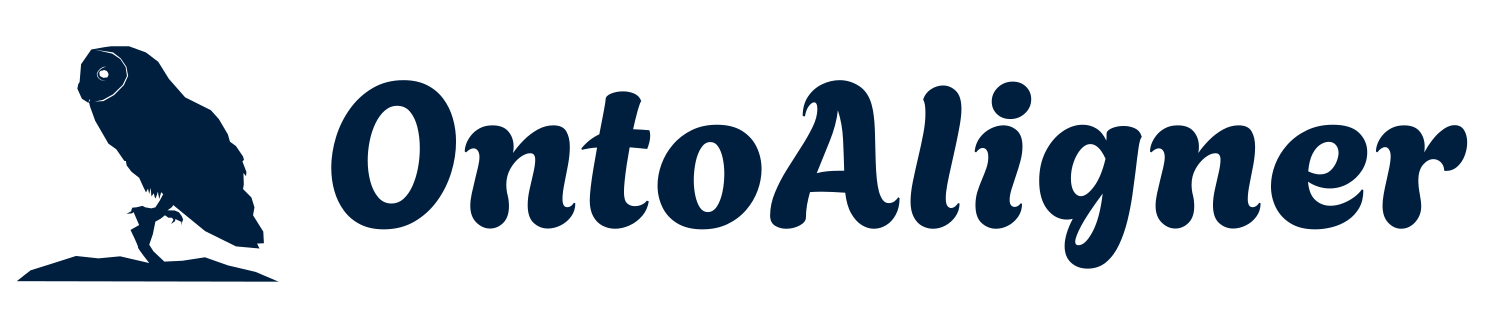}} % adjust numbers
  \end{picture}%
}

\makeatother

\insertlogo

\title{OntoAligner Meets Knowledge Graph Embedding Aligners}
% for Ontology Alignment
% \tnotetext[1]{You can use this document as the template for preparing your
  % publication. We recommend using the latest version of the ceurart style.}

% Add logo on the first page, top-right corner

\author[1]{Hamed {Babaei Giglou}}[%
orcid=0000-0003-3758-1454,
email=hamed.babaei@tib.eu]
\address[1]{TIB -- Leibniz Information Centre for Science and Technology, Hannover, Germany}

\author[1]{Jennifer D'Souza}[%
orcid=0000-0002-6616-9509,
email=jennifer.dsouza@tib.eu]

\author[1]{Sören Auer}[%
orcid=0000-0002-0698-2864,
email=auer@tib.eu]

\author[2]{Mahsa Sanaei}[%
orcid=0009-0008-2154-0627,
email=mahsasanaei97@ms.tabrizu.ac.ir]
\address[2]{University of Tabriz, Tabriz, Iran}

% %% Footnotes
% \cortext[1]{Corresponding author.}
% \fntext[1]{These authors contributed equally.}

\begin{abstract}
Ontology Alignment (OA) is essential for enabling semantic interoperability across heterogeneous knowledge systems. While recent advances have focused on large language models (LLMs) for capturing contextual semantics, this work revisits the underexplored potential of Knowledge Graph Embedding (KGE) models, which offer scalable, structure-aware representations well-suited to ontology-based tasks. Despite their effectiveness in link prediction, KGE methods remain underutilized in OA, with most prior work focusing narrowly on a few models. To address this gap, we reformulate OA as a link prediction problem over merged ontologies represented as RDF-style triples and develop a modular framework—integrated into the OntoAligner library—that supports 17 diverse KGE models. The system learns embeddings from a combined ontology and aligns entities by computing cosine similarity between their representations. We evaluate our approach using standard metrics across seven benchmark datasets spanning five domains: Anatomy, Biodiversity, Circular Economy, Material Science and Engineering, and Biomedical Machine Learning. Two key findings emerge: first, KGE models like ConvE and TransF consistently produce high-precision alignments, outperforming traditional systems in structure-rich and multi-relational domains; second, while their recall is moderate, this conservatism makes KGEs well-suited for scenarios demanding high-confidence mappings. Unlike LLM-based methods that excel at contextual reasoning, KGEs directly preserve and exploit ontology structure, offering a complementary and computationally efficient strategy. These results highlight the promise of embedding-based OA and open pathways for further work on hybrid models and adaptive strategies.
\end{abstract}



\maketitle

\section{Introduction}
Ontologies serve as formal, structured representations of knowledge within a specific domain. By axiomatizing concepts, relations, and properties, ontologies provide semantic richness that facilitates knowledge sharing, reasoning, and interoperability. Over the past decades, they have become a fundamental component of the Semantic Web, powering knowledge-intensive applications across domains ranging from E-commerce to Biomedical and Material Science. With the rise of artificial intelligence (AI), particularly in natural language processing (NLP), ontologies have increasingly been integrated into symbolic AI systems. The advent of large-scale transformer models, such as BERT~\cite{devlin-etal-2019-bert} and its successors, has accelerated this trend. These models have enabled new methods for semantic understanding, prompting researchers to revisit ontology-based approaches through the lens of deep learning. One such area that has gained substantial attention is Ontology Alignment (OA)~\cite{euzenat2007ontology} — the task of identifying correspondences between semantically equivalent entities across different ontologies.

Modern OA techniques predominantly leverage machine learning and embedding-based strategies to compute alignments. Among these, large language model (LLM) techniques have become especially popular due to their ability to encode contextual semantics~\cite{he2023exploring, qiang2023agent,hertling2023olala,babaei2024llms4om}. Despite this momentum, an important class of embedding methods — Knowledge Graph Embeddings (KGEs)~\cite{wang2017knowledge} — remains relatively underexplored in the context of OA. KGEs are techniques that convert the entities and relations in a knowledge graph into continuous low-dimensional vectors, allowing the graph’s semantic and structural information to be represented numerically. This transformation offers several advantages: it simplifies large and complex graphs, enables the use of machine learning algorithms, enhances search and recommendation systems by capturing semantic similarity, and helps uncover hidden patterns and relationships that are not easily detectable in the original symbolic form.

KGEs, such as TransE~\cite{bordes2013translating}, DistMult~\cite{yang2014embedding}, and ComplEx~\cite{trouillon2016complex}, are typically designed for single knowledge graphs, focusing on link prediction or taxonomy enrichment tasks within a single ontology. As a result, they are often perceived as less directly applicable to the cross-ontology matching objective of OA. However, advancements such as RDF2Vec~\cite{ristoski2016rdf2vec} and OWL2Vec~\cite{chen2021owl2vec} have demonstrated the potential of adapting KGE techniques for ontology-related tasks, including alignment, as is evident with OWL2VecOA~\cite{teymurova2024owl2vec4oa}. These models learn embeddings from RDF graphs or OWL ontologies by generating graph-based sequences, allowing them to capture semantic structure in a way that is more compatible with OA needs. Still, the majority of current work either focuses narrowly on a few KGE-based models or fails to harness the full spectrum of available KGE techniques in the alignment setting.  This gap highlights an important research opportunity: \textit{to systematically explore and adapt KGE models for OA, evaluate their comparative performance, and investigate hybrid models that integrate the strengths of both LLMs and KGEs}. By doing so, this work aims to advance the state of OA through KGE methodologies that are both semantically grounded and computationally scalable.

OA is the process of finding correspondences between semantically related entities from two ontologies. Formally, given two ontologies $O_{source} = (C_{s}, R_{s}, I_{s})$ and $O_{target} = (C_{t}, R_{t}, I_{t})$, where $C$, $R$, and $I$ denote sets of concepts, relations, and instances respectively, an alignment $A$ is a set of mappings $m = \langle e_{s}, e_{t}, r, \theta \rangle$ such that $e_{s} \in O_{s}$, $e_{t} \in O_{t}$, $r \in \{=, \subseteq, \supseteq, \equiv, \approx\} $ is a semantic relation, and $\theta \in [0, 1]$ represents the confidence score. A similarity function $\text{sim}(e_{s}, e_{t}): O_{source} \times O_{target} \to [0,1]$ is typically used to compute this score, and mappings are included in the alignment if $\text{sim}(e_{s}, e_{t}) \geq \tau$, where $\tau$ is a predefined similarity threshold. 

Additionally, an ontology can be represented as a set of triples in the form  $(h, r, t)$, where $h$ denotes the head entity, $r$ the relation, and $t$ the tail entity. In this work, to enable KGE models to learn meaningful representations from both $O_{source}$ and $O_{target}$, we combined their respective triples to construct a unified triple repository referred to as the triple factory $FA$. This integration serves as a foundational step toward a systematic framework for analyzing and exploring the behavior of KGEs within the context of OA. By formulating OA as a link prediction task over the merged ontologies, we implemented and evaluated a collection of KGE models specifically tailored for alignment objectives. This formulation allows the models to jointly learn latent representations of entities and relations across both ontologies, thereby capturing structural and semantic correspondences more effectively. The learned representations were subsequently utilized for equivalence-based alignment, enabling the identification of semantically equivalent concepts across ontologies.  Moreover, to make the collection entirely available for researchers and practitioners, we integrated our systematic approach to the \textit{OntoAligner}~\cite{babaei2025ontoaligner} -- a comprehensive and modular Python library dedicated to OA. 
% In addition to the modeling contributions, we conducted a detailed study of computational resource usage and efficiency trade-offs, providing practical insights into the scalability of KGE models for OA.  
% Moreover, to facilitate further research and real-world adoption, we integrated our complete pipeline into \textit{OntoAligner}~\cite{babaei2025ontoaligner}—a comprehensive, modular, and open-source Python library designed specifically for OA.

The remainder of this paper is organized as follows: \autoref{related-works} reviews related work; \autoref{methodology} details our proposed methodology with integration with the OntoAligner library; \autoref{results} presents the experimental results and analysis; and \autoref{conc-future-directions} concludes the paper with discussions and future research directions.

\section{Related Works}
\label{related-works}

Early work on graph alignment employs a graph embedding algorithm, and studies have shown that the great capability of the  KGE method is effective for aligning structurally similar ontologies and is more robust against alignment noise when dealing with graphs of different sizes and architectures~\cite{chen2025ontology}.

\cite{li2019multi} proposed a multi-view embedding model for biomedical OA using TransE and ConvE, demonstrating the utility of combining structural and semantic perspectives. Similarly, \cite{canastra2025systematic} conducted a systematic evaluation of KGEs for gene-disease association prediction, benchmarking models including TransD, TransE, TransH, DistMult, HolE, and ComplEx. More recent approaches have explored deeper models and alignment-specific enhancements. \cite{kim2022deep} introduced a security-aware, deep model-based entity alignment method incorporating MTransE, TransD, RotatE, ConvE, AlignE, AttrE, and GCN-a, tailored for edge-specific knowledge graphs. \cite{banerjee2024cross} tackled cross-lingual OA by leveraging both structural and semantic similarity via node2vec, GCN, RGCN, and TransE.

Contextual embeddings have also gained traction. \cite{chen2023contextual} combined traditional KGEs such as TransE, TransR, and DistMult with semantic embeddings like Word2Vec, Onto2Vec, OPA2Vec, and OWL2Vec to predict subsumption relations. Likewise, \cite{wang2023contextualized} presented LaKERMap, a contextualized structural self-supervised learning approach for ontology matching that employed TransE for inference tasks. Additionally, \cite{huang2023deep} used TransE, RotatE, and CompGCN for deep active alignment of knowledge graph entities and schemata. \cite{chen2022assertion} introduced an assertion and alignment correction framework using RDF2Vec, TransE, TransR, TransH, DistMult, and ComplEx.

Additional contributions include A-LIOn by \cite{alghamdi2022lion}, which used TransR to align ontologies through inconsistency-based negative sampling, and the AMD matcher \cite{wang2022amd}, which also adopted TransR for large-scale alignment scenarios. \cite{nguyen2022context} examined context-enriched models for aligning biomedical vocabularies using a range of methods, including TransE, TransR, RESCAL, DistMult, HolE, ComplEx, and ConvKB. In the scholarly knowledge domain, \cite{erten2024refining} refined the SemOpenAlex concept ontology using TransE, DistMult, and QuatE with SKOS-based constraints. \cite{ahmed2023nellie} introduced NELLIE, an open data linking system leveraging ComplEx embeddings for scalable entity linking. Finally, \cite{jain2022discovering} investigated fine-grained semantics in knowledge graph relations, further underscoring the breadth of KGE applications in ontology understanding and alignment.

\section{Methodology}
\label{methodology}

% \todo[inline,color=green!30]{The one-to-one mapping constraint adopted in the paper is unjustified, and it seems to me that KGE has the potential for identifying complex mappings where more than two entities across two ontologies are involved. This should be discussed in the paper, as after all in real-world mapping scenarios rather than OAEI, complex mappings can be common. Moreover, different mapping types other than one-to-one equivalence should be discussed in the paper, such as subsumption, as KGE should be capable of identifying them as well.}

The OntoAligner framework is composed of three key components: the \textit{parser}, \textit{encoder}, and \textit{aligner}. This modular design enables OntoAligner to serve as a hub for integrating diverse OA approaches. Building upon these design principles, we developed a framework that adheres to these foundational components, as illustrated in \autoref{framework}. In the following sections, we first introduce the overall alignment strategy, followed by an overview of the collection of KGE models integrated within this strategy.

\subsection{Graph Embedding Aligner}

\begin{figure}
    \centering
    \includegraphics[width=\linewidth]{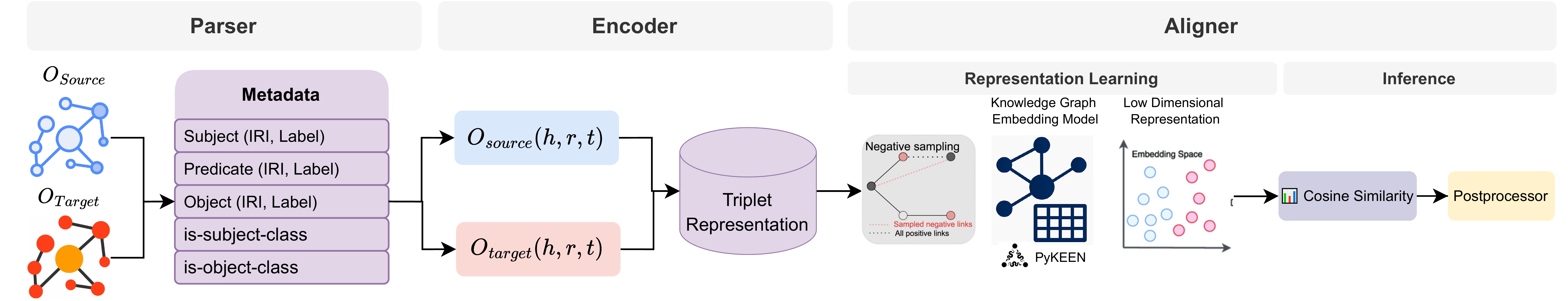}
    \caption{The architecture of the proposed framework, comprising three main stages: Parser, Encoder, and Aligner. The model ingests source and target ontologies, encodes them as unified triples, learns low-dimensional embeddings via a KGE model (trained in PyKEEN), and finally computes alignment predictions through cosine similarity.}
    \label{framework}
\end{figure}

The architecture of the \textit{Graph Embedding Aligner}, as illustrated in \autoref{framework}, follows a modular three-stage pipeline: Parser, Encoder, and Aligner. This design reflects the core components implemented in the code and provides a precise data flow from input ontologies to alignment predictions.

\paragraph{1) Parser.} The pipeline begins with parsing the source $O_{source}$ and target $O_{target}$ ontologies. Each ontology is decomposed into RDF-style metadata, consisting of Subject, Predicate, and Object, each annotated with its respective IRI and label. Moreover, a class membership metadata — indicating whether each entity plays a subject or object role in a class assertion is also attained. In the end, each ontology is represented by its own metadata.

\paragraph{2) Encoder.} In the encoding stage, the extracted triples from both ontologies are unified into a single triplet representation: a set of $O_{source}(h,r,t)$ and  $O_{target}(h,r,t)$ triples are obtained, where $h$, $r$, and $t$ are represented using natural language text rather than IRIs. Next, as a triplet representation, we unified both triples to form a triple factory $FA:= O_{source}(h,r,t) \cup O_{target}(h,r,t)$. One of the key advantages of this unified representation is that it enables the embedding model to automatically identify and learn shared structural and semantic patterns across both ontologies with high precision, thereby enhancing the quality and effectiveness of the alignment process.

\paragraph{3) Aligner.} The aligner component consists of two submodules. \textit{1) Representation Learning}, where a PyKEEN model embeds and trains each entity into a continuous vector space where semantically or structurally similar entities are positioned closely together. \textit{2) Inference}, which uses the learned embeddings to calculate cosine similarity between every $\langle e_{s} \in O_{source}, e_{t} \in O_{target}\rangle $ pair for ranking, and postprocessing.

% \todo[inline]{RW1: Unclear how the fine-tuning step is applied; not explained.}
\begin{itemize}
    \item \textsc{\textbf{Representation Learning}}. In this submodule, a KGE model is trained using a link prediction objective. The process begins with negative sampling, which augments the dataset by generating plausible but incorrect triples. This step helps the model learn to distinguish between valid and invalid relationships. The KGE model is then fine-tuned, and its resulting low-dimensional embeddings are used for alignment. Although negative sampling may seem to introduce noise into the OA process, it plays a critical role in improving the embedding model’s ability to distinguish between valid and invalid relations. This results in more robust and generalizable representations, which ultimately lead to more accurate alignment across heterogeneous ontologies. We utilize PyKEEN~\cite{ali2021pykeen} to support each stage of representation learning, as it offers a comprehensive suite of KGE models and tools.
    
    \item \textsc{\textbf{Inference}}.  In inference, once embeddings are learned, the embeddings $ \forall e \in O_{source}$ and $ \forall e \in O_{target}$ are extracted. Let $\mathbf{E}_{source} = [\mathbf{e}_s^{(1)}, \mathbf{e}_s^{(2)}, \dots, \mathbf{e}_s^{(n)}] \in \mathbb{R}^{n \times d}$ denote the matrix of L2-normalized embeddings for entities from the source ontology  $O_{source}$, and let $\mathbf{E}_{target} = [\mathbf{e}_t^{(1)}, \mathbf{e}_t^{(2)}, \dots, \mathbf{e}_t^{(m)}] \in \mathbb{R}^{m \times d}$ represent the normalized embeddings from the target ontology $O_{target}$, where $d$ is the embedding dimension. Each vector is normalized as $\| \mathbf{e} \|_2 = 1$. The similarity between entities is computed using cosine similarity. The similarity matrix $\mathbf{S} \in \mathbb{R}^{n \times m}$ is defined as $\mathbf{S} = \mathbf{E}_{source} \cdot \mathbf{E}_{target}^\top$. Where, each entry $S_{ij}$ corresponds to the cosine similarity between the $i$-th source and $j$-th target entities: $S_{ij} = \cos(\theta) = \mathbf{e}_s^{(i)} \cdot \mathbf{e}_t^{(j)} = \sum_{k=1}^{d} e_{s,k}^{(i)} \cdot e_{t,k}^{(j)}$. In the final step, for each source entity $i$, the target entity with the highest similarity score is selected: $j^* = \arg\max_{j} S_{ij}$. The alignment result is then given by: $\text{A}_i =  \langle e_s^{(i)}, e_t^{(j^*)}, \theta \rangle$. Where $\theta:= S_{ij^*}$ is the confidence score of the alignment.

    Once the alignment pairs are extracted, a post-processing step is applied to refine the results. First, a one-to-one cardinality constraint is enforced to ensure that each  $e_s \in O_{source}$ aligns with at most one $e_t \in O_{target}$, and vice versa. Then, a confidence-based filtering is performed by applying a similarity threshold $\theta \geq \tau$, where $\tau \in [0, 1]$ is a predefined cutoff. Alignment pairs with scores below this threshold are discarded to retain only the most confident and unambiguous matches.
\end{itemize}

\subsection{Knowledge Graph Embedding Collections}

\begin{table}[ht]
    \centering
    \caption{Overview of 17 KGE models supported in OntoAligner. }
    \label{kge-collection}
    \resizebox{\textwidth}{!}{ 
    \begin{tabular}{|l|p{13cm}|l|}
    \hline
    \textbf{KGE Model} & \textbf{Overview} & \textbf{Related Works} \\
    \hline
    \hline
      ConvE~\cite{dettmers2018convolutional}   & ConvE is a deep convolutional embedding model for link prediction that uses 2D convolutions over reshaped entity and relation embeddings to capture complex interaction patterns between them. & \cite{sun2022revisiting,wu2022leveraging,kalinowski2023scalable,li2019multi, kim2022deep}  \\
      \hline
      TransD~\cite{ji-etal-2015-knowledge} &  TransD (Translation on Dynamic Mapping Matrices) is a knowledge graph embedding model designed for link prediction and triple classification. It improves upon earlier translation-based models like TransE, TransH, and TransR by dynamically constructing relation-specific projection matrices using both entity and relation projection vectors. & \cite{canastra2025systematic,qiu2024earthquake, kim2022deep} \\
      \hline
      TransE~\cite{bordes2013translating} & TransE is a simple and scalable model for embedding knowledge graphs in low-dimensional vector spaces. It represents relationships as vector translations between head and tail entity embeddings. For a valid triple $(h,r,t)$, TransE enforces that $h+r \approx t$. The model is easy to train, has few parameters, and can handle very large datasets. & \cite{li2019multi, canastra2025systematic,banerjee2024cross,chen2023contextual,wang2023contextualized,huang2023deep,chen2022assertion, nguyen2022context,erten2024refining}\\
      \hline
      TransF~\cite{feng2016knowledge} & TransF is a flexible translation-based model for knowledge graph embedding. It improves on previous methods by allowing more adaptable translations to better handle complex relation types, like one-to-many or symmetric relations. Without increasing model complexity, TransF introduces a new scoring function and shows strong performance improvements in experiments.  & - \\
      \hline
      TransH~\cite{wang2014knowledge} & TransH improves knowledge graph embeddings by projecting entities onto relation-specific hyperplanes before translation. This allows it to handle complex relation types (e.g., one-to-many) better than TransE, while maintaining similar efficiency and scalability.  & \cite{canastra2025systematic,chen2022assertion}\\
      \hline
      TransR ~\cite{lin2015learning} & TransR enhances knowledge graph embeddings by projecting entities into relation-specific spaces before applying translations, allowing better modeling of diverse relational semantics than TransE and TransH.  & \cite{chen2023contextual,chen2022assertion,alghamdi2022lion,wang2022amd,nguyen2022context}\\  
      \hline
      DistMult~\cite{yang2014embedding} & DistMult is a simple bilinear embedding model for knowledge graphs that represents entities and relations as vectors, using matrix multiplication to capture relational semantics. It outperforms previous models like TransE in link prediction and enables effective logical rule mining.  & \cite{canastra2025systematic,chen2023contextual,chen2022assertion,nguyen2022context,erten2024refining,jain2022discovering}\\ 
      \hline
      ComplEx~\cite{trouillon2016complex} & ComplEx is a link prediction model that uses complex-valued embeddings to effectively capture both symmetric and antisymmetric relations. It relies on the Hermitian dot product, offering a simple yet powerful and scalable approach that outperforms existing models on standard benchmarks.  & \cite{canastra2025systematic,chen2022assertion,nguyen2022context,ahmed2023nellie}\\ 
      \hline
      HolE~\cite{nickel2016holographic} & HolE (Holographic Embeddings) is a knowledge graph embedding model that uses circular correlation to create compositional vector representations of entities and relations. It efficiently captures complex interactions while remaining scalable and easy to train.  & \cite{canastra2025systematic,nguyen2022context} \\  
      \hline
      RotatE~\cite{sun2019rotate} & RotatE is a knowledge graph embedding model that represents relations as rotations in complex space, enabling it to capture patterns like symmetry, inversion, and composition. It uses a novel self-adversarial negative sampling for efficient training and outperforms prior models on link prediction tasks.  & \cite{kim2022deep,huang2023deep} \\
      \hline
      SimplE~\cite{kazemi2018simple} & SimplE improves tensor factorization for knowledge graph link prediction by learning dependent embeddings for each entity, overcoming limitations of traditional methods. It offers interpretable, efficient embeddings, supports background knowledge, and achieves strong performance with proven full expressiveness.  & - \\
      \hline
      CrossE~\cite{zhang2019interaction} & CrossE is a knowledge graph embedding method that models bi-directional interactions between entities and relations by creating both general and triple-specific embeddings. It achieves state-of-the-art link prediction results on complex datasets and improves explainability by generating reliable paths to support its predictions.  & - \\
      \hline
      BoxE~\cite{abboud2020boxe} & BoxE is a knowledge base completion model that represents entities as points and relations as hyper-rectangles (boxes) in a spatial embedding. It overcomes key limitations of previous models by supporting logical rules, hierarchies, and higher-arity relations. & - \\
      \hline
      CompGCN~\cite{vashishth2019composition} & CompGCN is a graph convolutional network designed for multi-relational graphs that jointly learns embeddings for both nodes and relations. It uses entity-relation composition methods to efficiently handle many relations, generalizes existing multi-relational GCNs, and achieves strong results on tasks like node classification and link prediction.  & \cite{huang2023deep}\\
      \hline
      MuRE~\cite{balazevic2019multi} & MuRE proposes embedding multi-relational knowledge graphs in hyperbolic space using relation-specific transformations, better capturing multiple hierarchies. It outperforms Euclidean and other methods on link prediction, especially in low-dimensional settings.  & - \\
      \hline
      QuatE~\cite{zhang2019quaternion} & QuatE uses quaternion embeddings to model entities and relations in knowledge graphs, enabling expressive 4D rotations and compact interactions via the Hamilton product. It generalizes ComplEx with better geometric properties and effectively captures key relational patterns, achieving strong results on benchmark datasets.  & \cite{erten2024refining}\\
      \hline
      SE~\cite{bordes2011learning} & SE proposes a neural network-based approach to embed symbolic knowledge from diverse Knowledge Bases into a continuous vector space, preserving and enriching their structure.  & - \\
      \hline
    \end{tabular}
}
\end{table}

Based on a review of related works, we identified the \textbf{10} most frequently used KGE models that have been applied to OA from various perspectives—whether as standalone baseline models or as part of hybrid frameworks. These models consistently appear across numerous OA studies and knowledge engineering benchmarks. In addition to these top 10, we incorporated seven additional models that, while not as widely adopted in OA specifically, have shown strong performance and versatility in broader knowledge graph tasks such as link prediction, completion, and entity classification. \autoref{kge-collection} summarizes all 17 models, presenting their key characteristics and citations from recent literature.

The selection of these models was guided by several criteria. First, we prioritized diversity in modeling approaches, ensuring inclusion of translation-based (e.g., TransE, TransH), convolutional (e.g., ConvE), bilinear (e.g., DistMult), and neural graph-based methods (e.g., CompGCN). Second, we considered theoretical expressiveness and scalability, selecting models that are capable of handling large-scale ontologies with complex relational patterns. Third, we looked at empirical evidence from prior evaluations that demonstrated the effectiveness of these models across a range of domains~\cite{chen2025ontology}. Finally, compatibility with PyKEEN was an important practical consideration, allowing for unified implementation and experimentation within the OntoAligner framework. These selected models provide a representative and comprehensive foundation for experimenting with embedding-based OA strategies.

\subsection{Integration with OntoAligner}
\label{ontoaligner}
To enable embedding-based OA, we developed a specialized module called \textit{GraphEmbeddingAligner}. Built on top of the PyKEEN framework, this aligner harnesses KGE models to learn vector representations of entities from both source and target ontologies. OntoAligner currently supports \textbf{17} KGE models, all of which can be easily integrated through this module. A comprehensive usage guide is available at \url{http://ontoaligner.readthedocs.io/aligner/kge.html}. With a modular and extensible design, \texttt{GraphEmbeddingAligner} allows users to flexibly experiment with different KGE models and customize training configurations to suit various alignment tasks. An example demonstrating how to use KGE-based aligners within OntoAligner can be found at \url{https://github.com/sciknoworg/OntoAligner/blob/main/examples/kge.py}.

\section{Evaluations}
\label{results}
% \todo[inline]{RW1: It is concluded that the KGE aligner mostly achieves high precision but low recall, making it more suitable for high-confidence, low-risk integration tasks. However, this claimed superiority over traditional OA systems should be evaluated more deeply because precision/recall balance is always a matter of setting a threshold according to precision/recall preferences.}

% \todo[inline]{RW1: Further, the evaluation concludes that KGE aligners are relatively fast; however, the response time for the traditional OA system is not reported for comparison}

% \todo[inline]{RW1: Finally, it concludes that there is no universal threshold for a task or a domain, which would need adaptive thresholding. While it sounds reasonable, I missed evaluating the characteristics of the involved ontologies and their structure and complexity.}

% \todo[inline]{RW1: Evaluation results are summarised in Table 3 and in Figure 2. It would be helpful to provide detailed results (with performance measures in a table) for all KGE methods in the supplementary material.}

This section delves into empirically validating KGE models by employing precision, recall, and F1-score metrics. Experimental datasets and results are presented in the following.

\begin{table}
    \centering
    \caption{OAEI tracks and tasks statistics across source, target, and alignments.}
    \label{exp-datasets}
    \resizebox{\textwidth}{!}{ 
    \begin{tabular}{|l|l|r|r|r|}
        \hline
         \textbf{Track} & \textbf{Task} & \textbf{$O_{source}$ Triplets}& \textbf{$O_{target}$ Triplets} & \textbf{References} \\
        \hline
        \hline
        \textsc{Anatomy} & Mouse-Human &5,957 & 13,196 & 1,516 \\
        \hline
        \multirow{2}{*}{\textsc{Biodiv} -- Biodiversity \& Ecology} &  ENVO-SWEET & 16,237 & 17,182 & 805 \\
                                         &  FISH-ZOOPLANKTON & 440 & 173 & 15 \\
        \hline
        \textsc{CE}--Circular Economy & CEON - BiOnto & 1,896 & 1,307 & 18 \\
        \hline
        \textsc{MSE}--Material Science \& Engineering & MI - MatOnto &2,152 &2,740 & 302 \\
        \hline
        \multirow{2}{*}{\textsc{Bio-ML}--Biomedical Machine Learning} & NCIT-DOID & 22,559& 19,276 & 3,280 \\
                                         & OMIM-ORDO & 23,189 & 11,429 & 2,605 \\
        \hline
    \end{tabular}
    }
\end{table}

\subsection{Experimental Setups}

\paragraph{Evaluation Datasets OAEI Tracks and Tasks.}We carefully chose five tracks from the OAEI-2024 campaign~\cite{jimenez2025proceedings} spanning diverse domains for our experimental configurations. The statistics for seven datasets in five tracks are outlined in \autoref{exp-datasets}. The chosen tracks include: \textsc{Anatomy}~\cite{anatomy} (Mouse-Human), \textsc{biodiv} -- Biodiversity and Ecology~\cite{biodiversity} (two tasks), \textsc{CE} -- Circular Economy~\cite{blomqvist2023cross} (CEON-BiOnto), \textsc{MSE} -- Material Science and Engineering~\cite{mse}(MI-MatOnto), and \textsc{Bio-ML} -- Biomedical Machine Learning~\cite{bioml} (two tasks). These tracks were chosen to represent a range of dataset sizes (according to the number of triplets) and complexity levels, including: Small-scale tasks (\textsc{CE} track, \textsc{MSE} track, and FISH-ZOOPLANKTON from \textsc{Biodiv} track), Medium-scale tasks (\textsc{Anatomy}), and Large-scale tasks (\textsc{Bio-ML} track and the ENVO–SWEET task from the \textsc{Biodiv} track).  This selection ensures the evaluation covers both domain and size diversity, providing insights into the performance of KGEs.

\paragraph{KGE Hyperparameters and OS.} For fair comparison, we used CPU-based experimentation with an embedding dimension of 200, training epoch number of 20, train batch size of 64, evaluation batch size of 128, and a number of negative samples per positive sample of 5. Moreover, we used 10 core CPUs with a maximum memory of 80 GB for experimentation.

\subsection{Results}

The \autoref{tab:results} presents the best-performing KGE model for each of the seven OAEI benchmark tasks for all seven tasks of this study.  

% \begin{figure}
%     \centering
%     \includegraphics[width=\linewidth]{images/cpu_memory_plots.jpg}
%     \caption{Caption}
%     \label{fig:enter-label}
% \end{figure}

\begin{table}[]
    \centering
    \caption{Best-performing KGE models on seven OAEI-2024 benchmark tasks. The table reports performance metrics for the top KGE model per task, including threshold $\tau$, overlap between alignments and gold ($\cap$), alignment size ($A$), execution time ($\mathrm{T}$ in seconds), precision (Prec), recall (Rec), and F-Measurement (F). The final column lists the best-performing OA system for each task (with F-Measurement score in parentheses) for comparison.}
    \label{tab:results}
        \resizebox{\textwidth}{!}{ 
    \begin{tabular}{|l|l|r|r|r|r|r|r|r|l|}
         \hline
         \textbf{Task} & \textbf{KGE} & $\tau$ & $\cap$  & $A$ & $ \mathrm{T}$ & \textbf{Prec} & \textbf{Rec} & \textbf{F} & \textbf{Best Performer}\\
         \hline
         \hline
         Mouse-Human & DistMult &  0.34 & 1047 & 1069 & 70.3 & 97.9 & 69.0 & 81.0 & Matcha~\cite{cotovio2024matcha} (94.1)\\ % 42s (2023 -> 54)
        \hline
        FISH-ZOOPLANKTON & TransF &  0.25 & 9 & 9 & 4 & 100 & 60.0 & \textbf{74.9} & LogMapLt~\cite{jimenez2011logmap} (64.4) \\ %0
        ENVO-SWEET & ConvE &  0.6 & 327 & 367 & 855 & 89.1 & 40.6 & 55.8 &‌LogMap~\cite{jimenez2011logmap} (71.3) \\ %21
        \hline
        CEON-BiOnto & ConvE &  0.35 & 10 & 17 & 80 & 58.8 & 55.5 & \textbf{57.1} & Matcha~\cite{cotovio2024matcha} (47.8)\\ 
        \hline
        MI-MatOnto & TransD &  0.39 & 32 & 37 & 29 & 86.4 & 10.5 & 18.8 & LogMap~\cite{jimenez2011logmap} (32.0)\\ 
        \hline
        OMIM-ORDO & ConvE &  0.43 & 538 & 772 & 871& 69.6 & 20.6 & 31.8 & BERTMap\textsuperscript{\textdaggerdbl}~\cite{jimenez2011logmap} (64.6) \\ 
        NCIT-DOID & SE &  0.43 & 1751 & 2537 & 1584 & 69.0 & 53.3 & 60.2 & HybridOM*~\cite{totoian2024hybridom} (91.8)\\
         \hline
    \end{tabular}
    }
\end{table}

\subsubsection{Domain Specific Analysis}
We have explored KGE models across five domains, and the results presented in \autoref{tab:results} present interesting findings per domain, such as:

\begin{itemize}
    % https://oaei.ontologymatching.org/2024/results/anatomy/index.html
    % Mouse-Human
    % - In terms of precision, our technique goes in the second place!
     \item \textbf{Anatomy Track.} In the Mouse-Human task, the \textit{DistMult} model reached a high precision of 97.9—ranking second among all methods of OAEI-2024\footnote{\url{https://oaei.ontologymatching.org/2024/results/anatomy/index.html}}—though its recall (69.0\%) brought the overall F-Measure to 81.0, behind Matcha aligner~\cite{cotovio2024matcha} with F-Measure of 94.1. % This shows promise in precision-focused applications, though improvements in recall are needed.

    % https://oaei.ontologymatching.org/2024/results/biodiv/index.html
    % FISH-ZOOPLANKTON 
    % * kge stood out than the best performer of 2024 (LogMapLt)
    % * https://oaei.ontologymatching.org/2024/results/biodiv/index.html
    % * it is a high-precisin matcher
    % * response time is almost zero for LogMapLt but here it is 4.55
    % SWEET-ENVO
    % - high precision that LogMap (which is the best performer)
    % - another performer that stood out in precision is LogMapLt (with precision of 80.3), but we are 89.1
    \item \textbf{Biodiversity and Ecology Track.} In the \textit{FISH-ZOOPLANKTON} task of \textsc{Biodiv}, KGE models -- \textit{TrasnF} aligner -- showed notable superiority. The TransF model achieved a perfect precision of 100 and an F-Measure of \textbf{74.9\%}, significantly outperforming LogMapLt~\cite{jimenez2011logmap}—the best OAEI-2024 system at this task\footnote{\url{https://oaei.ontologymatching.org/2024/results/biodiv/index.html}}—with an F1 of 64.4. Despite LogMapLt’s nearly perfect execution time, the KGE method required only 4 seconds, maintaining efficiency while outperforming the state-of-the-art system. In the \textit{SWEET-ENVO} task, the LogMap~\cite{jimenez2011logmap} remains the best OAEI system by F-Measure of 71.3\%. However, \textit{ConvE} aligner exhibited a higher precision of 89.1\% than all systems within OAEI 2024, including LogMapLt with a precision of 80.3\%. % This suggests that KGE models may offer greater reliability in tasks where precision is critical.

    % https://oaei.ontologymatching.org/2024/results/ce/index.html
    % CEON-BioOnto
    % * best performer is Matcha but we again are better than Matcha (also top precisioner) but we are top performer here!
    \item \textbf{Circular Economy Track.} For the \textit{CEON-BiOnto} task from this track, the \textit{ConvE} aligner achieved the best overall F-Measure of \textbf{57.1}\%, outperforming Matcha~\cite{cotovio2024matcha} with an F-Measure of 47.8\%, the top performer in the official OAEI-2024\footnote{\url{https://oaei.ontologymatching.org/2024/results/ce/index.html}} results. 

    % The model also attained the highest precision, demonstrating its effectiveness in specialized ontology domains.
    % https://github.com/EngyNasr/MSE-Benchmark/tree/main/Results/OAEI2023
    % MI-MatOnto (2nd case)
    % * LogMap	67	p=0.881	 r=0.195	f1=0.320	time=06
    % * in terms of precision we are 1.5\% fall short.
    % - but this precions is the second place!
    \item \textbf{Material Science and Engineering Track.} The \textit{MI-MatOnto} task within this track, showed a poor F-Measure of 18.8\% using  \textit{TransD} aligner. However, the maintained precision of 86.4\% places this model's performance close to the LogMap system (from OAEI-2023\footnote{\url{https://github.com/EngyNasr/MSE-Benchmark/tree/main/Results/OAEI2023}}) with a precision of 88.1\%.

    % TransD performed poorly in recall (10.5), limiting its F1 to 18.8. However, it maintained strong precision (86.4), placing it close to LogMap’s 88.1. This reflects a common KGE behavior of favoring precision at the cost of coverage.
    % https://krr-oxford.github.io/OAEI-Bio-ML/2024/index.html#results
    % OMIM-ORDO
    % (equivelance - unsupervised)
    % - did not work well even in terms of precion
    % - NCIT-DOID also didn't work well!
    \item \textbf{Biomedical Machine Learning Track.} For this track, the KGE models did not perform well in terms of precision and F-Measure. For the \textit{OMIM-ORDO} task, \textit{ConvE} aligner reached only F-Measure of 31.8\%, far behind BERTMap\textsuperscript{\textdaggerdbl}~\cite{jimenez2011logmap} with F-Measure of 64.6\% (OAEI-2024\footnote{\url{https://krr-oxford.github.io/OAEI-Bio-ML/2024/index.html}} system performance). Even the precision of 69.6\% fell short, indicating limitations of unsupervised KGE methods for complex disease-related alignments. For another task of this track, specifically \textit{NCIT-DOID} task, it showed \textit{SE} aligner performance of 60.2\% in terms of F-Measure, significantly underperforming compared to HybridOM* ~\cite{totoian2024hybridom} with F-Measure of 91.8\%. Precision of 69.0\% was still promising, but the gap in recall of 53.3\% limited its overall effectiveness.
\end{itemize}

KGE models, particularly \textit{ConvE} and \textit{TransF}, demonstrated competitive or even superior performance on two tasks—\textit{FISH-ZOOPLANKTON} and \textit{CEON-BiOnto}—in terms of F-Measure. In other tasks, particularly OMIM-ORDO and NCIT-DOID, performance lagged behind traditional or supervised OA systems. Overall, KGE methods tend to produce \textbf{high-precision} alignments with lower recall, suggesting their suitability for applications requiring conservative, high-confidence mappings.

\begin{figure}
    \centering
    \includegraphics[width=\linewidth]{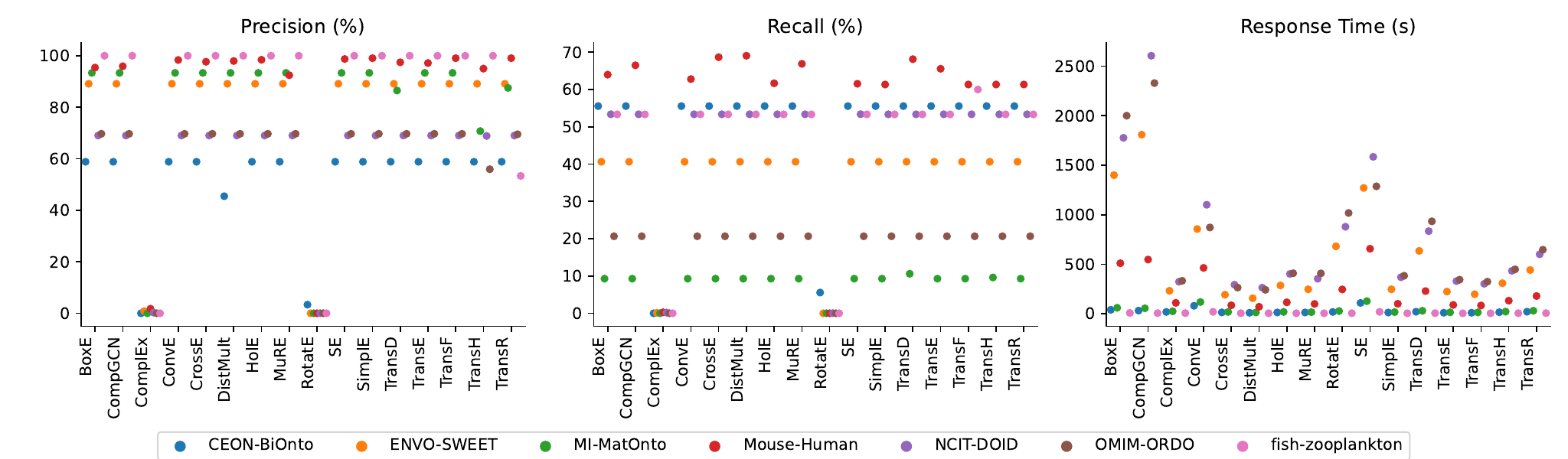}
    \caption{Column 1 and 2: Precision and recall analysis across aligners and tasks. Column 3: Representation learning and inference time response analysis.}
    \label{fig:precision-recall-response-time}
\end{figure}

\subsubsection{Empirical Trends in KGE Aligners}
% \todo[inline,color=green!30]{Lastly, the title of section 4.2.2 does not seem fit, and I don't see any "general patterns" revealed.}
% The statistics for seven datasets in five tracks are outlined in \autoref{exp-datasets}. In addition, comprehensive task-specific results for each KGE model are presented in \autoref{app:detailed-results}. 
% The results highlight the differences in aligner effectiveness across diverse domains, ranging from biomedical (e.g., OMIM–ORDO, NCIT–DOID) to environmental (e.g., ENVO–SWEET, FISH–ZOOPLANKTON) and cross-species settings (e.g., Mouse–Human). While some models such as TransE, TransD, and DistMult demonstrate consistently strong trade-offs between precision and recall, others like ComplEx and RotatE show unstable behavior across tasks, often failing to produce meaningful alignments. Furthermore, execution time varies substantially among methods, with graph convolutional models (e.g., CompGCN) being more computationally demanding than simpler translational approaches. Overall, these tables provide a fine-grained view of model behavior, complementing the aggregated results reported in the main paper.

The analysis of the summary of results presented in \autoref{tab:results} across seven benchmark tasks and five domains reveals consistent empirical trends in KGE aligners' behavior. While performance varies across tasks and domains, certain trends emerge regarding precision, recall, alignment size, execution time, and model-task interactions. These observations provide insight into the operational characteristics and efficiency of KGE aligners and highlight considerations for their application in ontology alignment tasks

% These findings offer insight into the behavioral characteristics, operational efficiencies, and model-task interactions of KGE approaches in OA.

\paragraph{Behavioral Characteristics.} Across all tasks, the top-performing KGE models exhibited notably high precision scores—often exceeding 85\%—even in tasks where recall and F-Measure were relatively poor, e.g., TransD in MI-MatOnto (Precision 86.4\% but Recall 10.5\%), ConvE in ENVO-SWEET (Precision 89.1\% vs. Recall 40.6\%), or DistMult in Mouse-Human: (Precision 97.9\% vs. Recall 69.0\%). More detailed analysis is presented in \autoref{fig:precision-recall-response-time} (Precision and Recall columns). This shows that KGE aligners are conservative aligners that prioritize correctness over completeness, making them well-suited for high-confidence, low-risk integration tasks where false positives are costly.

Moreover, tasks like OMIM-ORDO and NCIT-DOID, which have large-scale sizes in terms of references, also had large-scale alignments ($A=772$ and $A = 2537$ respectively). Yet, their performance (F-Measures of 31.8\% and 60.2\%, respectively) remained moderate compared to tasks with smaller alignments (e.g., FISH-ZOOPLANKTON with only nine alignments and F-Measures = 74.9\%, where total references is 9 -- see \autoref{exp-datasets} for total references). This means that predicted alignment volume does not directly translate to quality, especially in semantically dense or noisy domains. KGE models may overgenerate candidates in large ontologies unless appropriately constrained.

\begin{figure}
    \centering
    \includegraphics[width=0.7\linewidth]{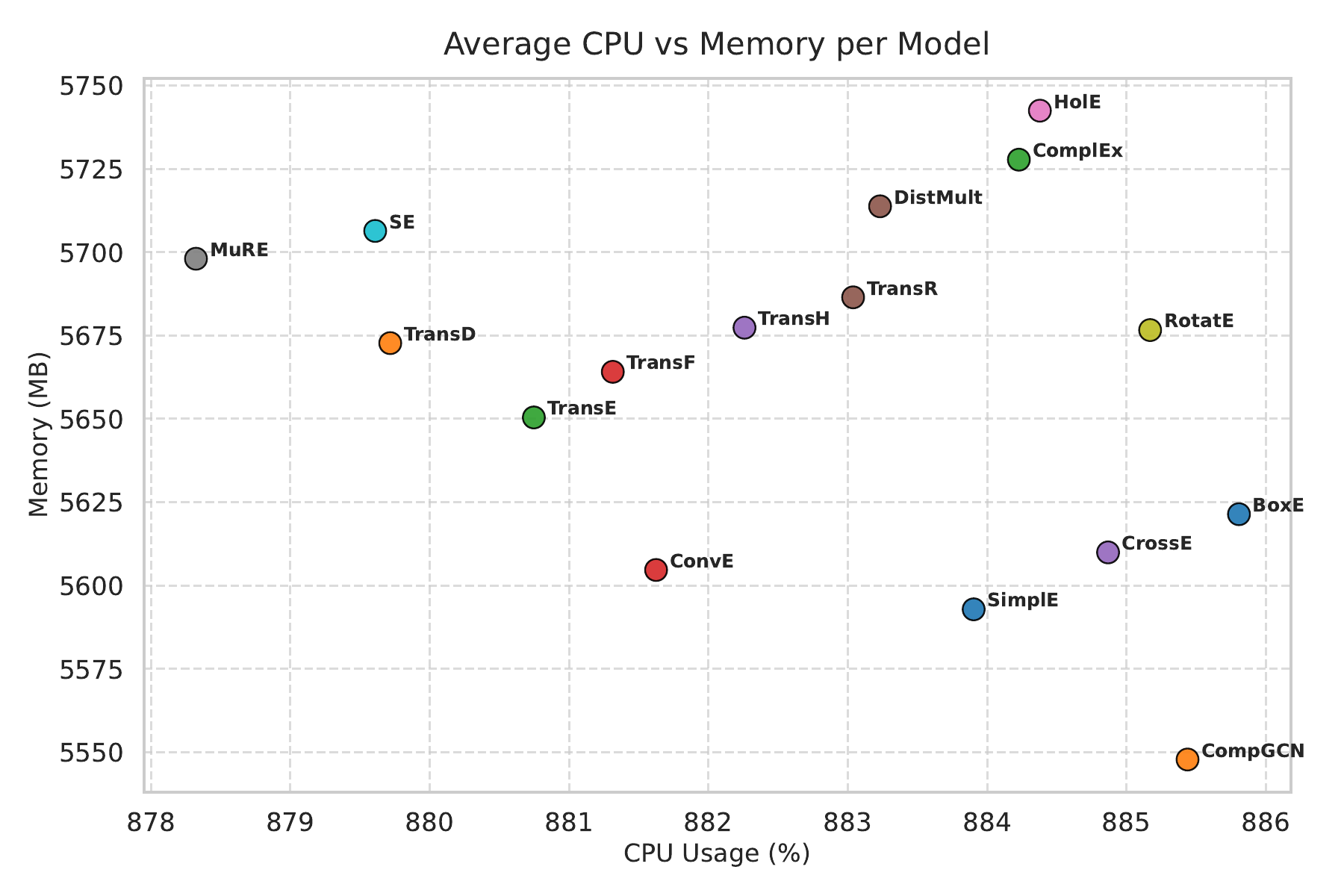}
    \caption{Scatter plot showing the average CPU utilization (\%) against memory consumption (MB) for each KGE aligner. Each color represents a distinct model.  This visualization highlights the relative computational efficiency and resource demands of the models.}
    \label{fig:cpu-vs-memory}
\end{figure}

% \paragraph{Operational Efficiency.} The majority of KGE aligners completed their tasks--representation learning and inference--within under 100 seconds (4 tasks according to \autoref{tab:results}), including those with hundreds or thousands of candidate alignments, such as Mouse-Human. The response times of KGE aligners are presented in \autoref{fig:precision-recall-response-time} (Response Time column). KGE-based methods are computationally efficient, making them viable for scalable or on-demand OA, especially in dynamic systems or real-time services.
\paragraph{Operational Efficiency.} The majority of KGE aligners completed their tasks—including representation learning and inference—within 100 seconds for most benchmark tasks (\autoref{tab:results}), even when handling hundreds or thousands of candidate alignments, such as in Mouse-Human. \autoref{fig:cpu-vs-memory} illustrates the CPU versus memory usage of the models, averaged across all tasks. Most aligners utilize 80–90\% of available CPU cores and consume over 6GB of memory, highlighting that KGE-based methods are computationally efficient and capable of supporting scalable or on-demand ontology alignment, including in dynamic or real-time systems.

\paragraph{Model-Task Interactions.} The optimal similarity threshold $\tau$ for alignment significantly varies across tasks, suggesting that no universal threshold works across domains. This indicates that KGE aligners require task-specific or domain-specific calibration, particularly around the similarity threshold. Auto-tuning or adaptive thresholding could significantly improve F-Measure in future iterations.

Nevertheless, certain KGE aligners seem particularly effective in specific domains. ConvE aligner performs best in comparison to its own companion KGE aligners in multi-relational tasks (ENVO-SWEET, CEON-BiOnto, OMIM-ORDO). TransF aligner excels in structure-rich, less ambiguous domains (FISH-ZOOPLANKTON). DistMult works well in clean, hierarchical taxonomies (Mouse-Human). SE aligner provides a better balance of precision and recall in larger biomedical terminologies (NCIT-DOID). Future systems could auto-select the KGE aligner based on ontology metadata (e.g., size, depth, domain) to optimize performance per task or do ensemble learning.

\section{Conclusions and Future Directions}
\label{conc-future-directions}

In this paper, we have systematically explored the application of Knowledge Graph Embeddings within Ontology Alignment tasks. Our comprehensive framework, integrated within the OntoAligner toolkit, leverages a collection of 17 prominent KGE models.  Through empirical evaluations on seven benchmark tasks across diverse domains such as Anatomy, Biodiversity, Circular Economy, Material Science, and Biomedical Machine Learning, we identified several key findings. KGE-based aligners generally produce alignments characterized by high precision but moderate recall, indicating their suitability for conservative, high-confidence ontology matching scenarios.

Notably, ConvE and TransF emerged as particularly effective models, demonstrating superior performance in multi-relational and structure-rich tasks, respectively. Nevertheless, our analysis also revealed limitations of KGE methods in complex biomedical alignments, highlighting the need for improved techniques or hybrid approaches in these domains.

Future research could address several promising directions:  
\begin{itemize}
    \item \textit{Adaptive Thresholding and Calibration:} Our results indicated no universal similarity threshold applicable across diverse ontology alignment tasks. Developing adaptive thresholding strategies that dynamically calibrate based on ontology metadata could enhance model flexibility and performance.
    \item \textit{Hybrid Models:} Integrating KGEs with LLMs or other contextual embedding approaches could leverage complementary strengths, potentially improving alignment accuracy in complex, context-rich domains.
    \item \textit{Domain-specific Enhancements:} Given the domain-dependent effectiveness observed, tailoring KGE methodologies to specific ontological structures or leveraging metadata-driven model selection and ensemble strategies may provide meaningful gains.
\end{itemize}

In conclusion, embedding-based ontology alignment presents a powerful yet still evolving paradigm. 
Addressing these future directions will not only advance the state-of-the-art in ontology alignment but also extend the practical utility of KGEs across a broader range of semantic web and knowledge-intensive applications.

\begin{acknowledgments}
This work is jointly supported by the \href{https://scinext-project.github.io/}{SCINEXT project} (BMFTR, German Federal Ministry of Research, Technology and Space, Grant ID: 01lS22070), the KISSKI AI Service Center (BMFTR, Grant ID: 01IS22093C), and the \href{https://www.nfdi4datascience.de/}{NFDI4DataScience initiative} (DFG, German Research Foundation, Grant ID: 460234259).

\end{acknowledgments}

%% The declaration on generative AI comes in effect
%% https://ceur-ws.org/GenAI/Policy.html

\section*{Declaration on Generative AI}
In preparing this manuscript, generative AI tools—specifically ChatGPT—were used solely for: grammar checking, spelling check, and the readability of some sentences.  All suggested changes were carefully reviewed and adapted by the authors to ensure accuracy and appropriateness. The scientific content, research design, analysis, and conclusions were developed and verified exclusively by the authors without AI involvement. The use of ChatGPT was limited to enhancing the presentation of the work.
\bibliography{mybib}

\end{document}